\documentclass[sigconf,natbib=false]{acmart}

\AtBeginDocument{%
  }

\copyrightyear{2023}
\acmYear{2023}
\setcopyright{rightsretained}
\acmConference[MADiMa '23] {Proceedings of the 8th International Workshop on Multimedia Assisted Dietary Management}{October 29, 2023}{Ottawa, ON, Canada.}
\acmBooktitle{Proceedings of the 8th International Workshop on Multimedia Assisted Dietary Management (MADiMa '23), October 29, 2023, Ottawa, ON, Canada}
\acmPrice{}
\acmISBN{979-8-4007-0284-6/23/10}
\acmDOI{10.1145/3607828.3617796}
\RequirePackage[
  datamodel=acmdatamodel,
  style=acmnumeric,
  ]{biblatex}

\begin{document}

\title{Diffusion Model with Clustering-based \\ Conditioning for Food Image Generation}


\author{Yue Han}
\affiliation{%
  \institution{Purdue University}
  \city{West Lafayette}
  \state{IN}
  \country{USA}}
\email{han380@purdue.edu}

\author{Jiangpeng He}
\affiliation{%
  \institution{Purdue University}
  \city{West Lafayette}
  \state{IN}
  \country{USA}}
\email{he416@purdue.edu}

\author{Mridul Gupta}
\affiliation{%
  \institution{Purdue University}
  \city{West Lafayette}
  \state{IN}
  \country{USA}}
\email{gupta431@purdue.edu}

\author{Edward J. Delp}
\affiliation{%
  \institution{Purdue University}
  \city{West Lafayette}
  \state{IN}
  \country{USA}}
\email{ace@purdue.edu}

\author{Fengqing Zhu}
\affiliation{%
  \institution{Purdue University}
  \city{West Lafayette}
  \state{IN}
  \country{USA}}
\email{zhu0@purdue.edu}

\begin{abstract}
Image-based dietary assessment serves as an efficient and accurate solution for recording and analyzing nutrition intake using eating occasion images as input. 
Deep learning-based techniques are commonly used to perform image analysis such as food classification, segmentation, and portion size estimation, which rely on large amounts of food images with annotations for training. However, such data dependency poses significant barriers to real-world applications, because acquiring a substantial, diverse, and balanced set of food images can be challenging.
One potential solution is to use synthetic food images for data augmentation. Although existing work has explored the use of generative adversarial networks (GAN) based structures for generation, the quality of synthetic food images still remains subpar. In addition, while diffusion-based generative models have shown promising results for general image generation tasks, the generation of food images can be challenging due to the substantial intra-class variance.
In this paper, we investigate the generation of synthetic food images based on the conditional diffusion model and propose an effective clustering-based training framework, named ClusDiff, for generating high-quality and representative food images. 
The proposed method is evaluated on the Food-101 dataset and shows improved performance when compared with existing image generation works.
We also demonstrate that the synthetic food images generated by ClusDiff can help address the severe class imbalance issue in long-tailed food classification using the VFN-LT dataset.
\end{abstract}

\ccsdesc[500]{Computing methodologies~Computer vision tasks}
\ccsdesc[300]{Applied computing~Life and medical sciences}
\ccsdesc[500]{Computing methodologies~Machine learning}

\keywords{Food Image Generation, Diffusion Model, Food Classification, Image-based Dietary Assessment}


\maketitle

\section{Introduction}
Dietary intake has a profound impact on personal health and disease. 
A healthy diet is an essential aspect of keeping lifelong well-being for current and future generations across the lifespan. 
Dietary factors are crucial in determining the risk of developing obesity, heart disease, diabetes, and even cancer \cite{key2020diet}.
Dietary assessment has been utilized as a fundamental approach for understanding diet's effects on human health.
Traditional methods of dietary assessment are mainly based on self-reporting, which makes it difficult to obtain an accurate dietary assessment of individuals due to both random and systematic measurement errors \cite{bailey2021overview}.
Recently, image-based dietary assessment has been widely used to assist researchers and participants in recording their dietary intake with high accuracy and efficiency~\cite{boushey2017new}. Methods such as FoodLog~\cite{food_log}, DietCam~\cite{diet_cam}, and FoodCam~\cite{8} allow participants to capture their food intake using a mobile device, then the captured images are analyzed by trained dietitians to estimate the nutrient composition. 
To further reduce the human effort in analyzing the captured eating occasion images, several recent approaches~\cite{shao2021_ibdasystem} automate this process by using deep learning-based techniques to recognize the food~\cite{he2020multitask, he2021end, Mao2021ImprovingDA} and estimate the portion size \cite{fang-icip2018, fang2019end, shao2021towards} in the captured images. Besides using mobile devices to capture the food intake image, passive capture of eating occasion images utilizes wearable cameras assisted by on-device sensors to collect dietary intake data \cite{sun2022improved, han2021improving}. 

The image-based dietary assessment methods reduce the time and labor required for nutrient analysis and allow real-time feedback to the participants. 
The major challenge with deep learning-based approaches is that the performance is heavily dependent on the quality and the quantity of the datasets used for training the models.
It is difficult to obtain a large number of accurately annotated food images in real-world scenarios. Furthermore, food images collected in daily life usually suffer from data imbalance issues due to the nature that some foods are consumed more frequently than others~\cite{he2022long, he2023singlestage}.
One way to increase the size of the food image dataset is by using data augmentation methods to create more training samples. Traditional data augmentation methods utilize basic image manipulations including color space transformations, random erasing, geometric transformations, and kernel filtering\cite{castro2018elastic, montserrat2017training}.  
However, these augmentation methods do not work well when the training data is limited and can result in overfitting \cite{shorten2019survey}. With the development of Generative Adversarial Networks (GANs), generating synthetic images from GANs for data augmentation is widely used for various deep learning methods ~\cite{deepsynth, han2023ensemble}. 

Generative adversarial networks (GANs) have shown great potential in generating realistic synthetic images~\cite{goodfellow2014generative}. Moreover, conditional generative models \cite{mirza2014conditional} can generate images for a specific food class.
Previous works \cite{ito2018food, horita2019unseen, fu2023conditional} for synthetic food image generation have typically focused on generating images of only one or a limited number of food classes, primarily due to the challenge of creating a dataset with a diverse variety of foods.
Therefore, the limitation on scale of the current food image generation methods narrows the application of using them as data augmentation methods.
Diffusion models have recently become the hottest research topic in the area of generative models, showing great capability in image generation \cite{rombach2022high}, super-resolution \cite{gao2023implicit}, image modification \cite{kawar2023imagic}, and image inpainting \cite{lugmayr2022repaint}. However, there is no existing work utilizing the diffusion model for food image generation. One of the major challenges of applying the diffusion model is that the food data usually suffers from high intra-class variance due to the differences in recipes or cooking methods with the same ingredient \cite{jiang2020deepfood}. Therefore, it is relatively easy to have the mode collapse during the training of the image generation model. 
Mode collapse refers to the condition when the model only generates a limited set of examples instead of exploring the entire training data distribution \cite{thanh2020catastrophic}.

In this paper, we first explore the performance of latent diffusion methods on food image generation by fine-tuning a stable diffusion model\cite{rombach2022high} on the Food-101 dataset\cite{bossard2014food} and compare results to the images generated using one of the latest GAN-based methods, StyleGAN3 \cite{karras2021alias}. 
We propose a novel clustering-based training framework, ClusDiff, to solve the high intra-class variance issue when training the food image generation model.
For training ClusDiff, food images within each class are clustered as sub-classes, and then the model is trained with these sub-classes.
When generating images for a specific class of food, ClusDiff is conditioned on its sub-class labels, where the occurrence of the labels follows the distribution of sub-classes within a class. 
We evaluate ClusDiff on the Food-101 dataset and demonstrate that our proposed method achieves better food image generation performance in terms of Frechet Inception Distance Metric (FID)  \cite{heusel2017gans} as compared to the baseline, which is a fine-tuned stable diffusion model. 
We also explore the benefit of using synthetic images for data augmentation by providing a case study that uses synthetic food images generated by ClusDiff to address the class-imbalance issue in the VFN-LT dataset~\cite{he2022long} for long-tailed food image classification. Overall, the contribution of this work can be summarized into the following:
\begin{itemize}
    \item This work represents the first investigation and application of the diffusion model to food image generation.
    \item We propose ClusDiff, a diffusion model with clustering-based conditioning, and show that the proposed method generates better food images than existing methods.
    \item We demonstrate synthetic food images generated by ClusDiff can be used to address the data imbalance issue in food classification through a case study.
\end{itemize}

\section{Related Work}
\subsection{Image-based Dietary Assessment}

With recent advances in machine learning and modern computer vision, image-based dietary assessment methods have been introduced to automatically record and analyze participants' food intake. The most important aspect of image-based dietary assessment is to detect and classify the food and beverages in the images captured by participants.
DiaWear\cite{shroff2008wearable} is one of the pioneering works that uses a neural network to perform context-aware food recognition on food images captured by mobile devices. Deep learning-based detection methods, such as Faster R-CNN\cite{girshick2017faster} and YOLO\cite{redmon2016you} are frequently used for food detection and have shown good results when trained on a large number of ground truth images\cite{han2021improving, ege2018multi, jubayer2021detection}. 
The most recent work focuses on estimating food portion size or energy directly from the eating occasion images. In \cite{fang2016comparison}, a geometric model is introduced to estimate the food portion size. Conditional GAN~\cite{mirza2014conditional} is used in~\cite{fang-icip2018} to generate the energy distribution map with an end-to-end regression framework for food energy estimation~\cite{fang2019end}, which is further improved in~\cite{shao2021towards} by integrating information from RGB images and in~\cite{shao2023endtoend} by leveraging the 3D reconstruction. By combining these components, systems like goFOOD\cite{lu2020gofoodtm} and Technology-Assisted Dietary Assessment (TADA)\cite{shao2021_ibdasystem} can provide an end-to-end automatic nutrient estimation for a meal.

Despite the promising advancements in image-based dietary assessment methods, the performance of deep learning-based methods relies heavily on the quality and quantity of the datasets. Training a model for food recognition or portion estimation requires a large, diverse, and accurately annotated dataset.
Nevertheless, the food images we collect from real-world scenarios often exhibit stark differences from established food image datasets such as Food2K~\cite{Food2K}, Food-101~\cite{bossard2014food}, and UEC-256~\cite{uec-256}. 
This discrepancy primarily stems from inherent limitations associated with real-world data, such as variability, inconsistent sizes, and potential inaccuracies in annotations. While various food recognition systems have been developed to target real-world scenarios, such as continual learning~\cite{ILIO, He_2021_ICCVW, he2023longtailed, raghavan2023online, he2022_expfree}, few-shot learning~\cite{food_fewshot}, and long-tailed classification~\cite{gao2022dynamic_LTingredient, he2022long, he2023singlestage}, they usually necessitate more sophisticated training regimes. 
This amplification in complexity invariably leads to a surge in computational requirements, which are typically constrained in real-world situations. 
A potential solution to mitigate the problem of data scarcity in the real world without requiring any changes to the existing image-based dietary assessment system is to use synthetic food images for the training process. 
However, generating synthetic food images could be challenging due to inter-class similarity and intra-class variability. 
In this work, we investigate the application of the stable diffusion model\cite{rombach2022high} for synthetic food image generation.
We propose a novel framework for training a stable diffusion model to generate more diverse synthetic food images and demonstrate that these images can be used to address the data imbalance issue in tasks involving long-tailed food classification.

\begin{figure*}
  \centering
     \includegraphics[width=0.98\linewidth]{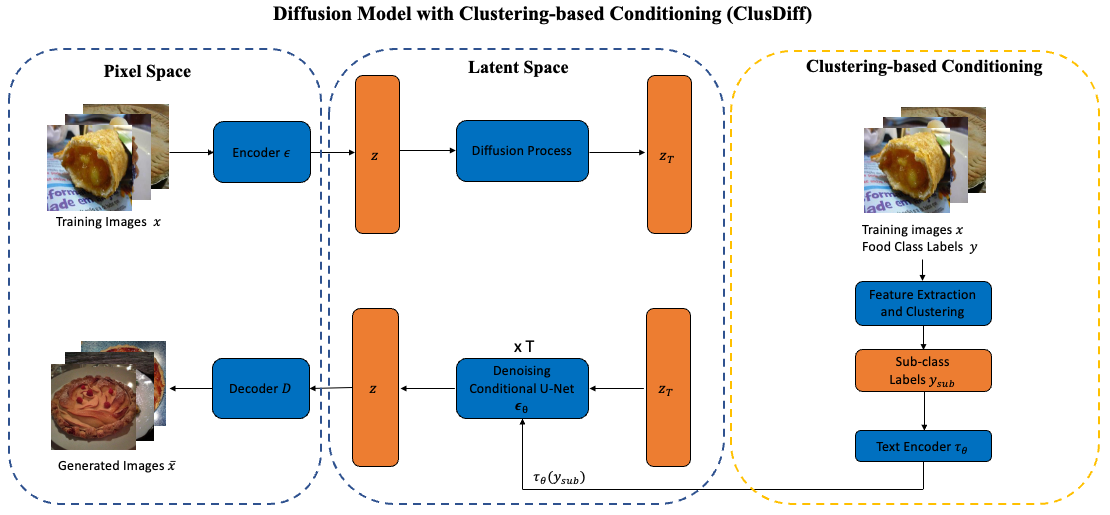}
  \caption{The block diagram of the proposed clustering-based conditional training strategy for latent diffusion model (ClusDiff). The blue blocks indicate the network structure, and the red blocks indicate data passing through the network.}
  \label{fig:system_bd}
\end{figure*}

\subsection{Synthetic Image Generation}
One of the main challenges of supervised machine learning or deep learning methods is the requirement for a large amount of annotated training samples, and the performance heavily relies on the quality of the dataset. 
One way to address this issue is by generating high-quality image data to augment the training dataset.
For a long time, Generative Adversarial Networks (GANs) have been the most impactful methods and have shown remarkable results for synthetic image generation. 
These synthetic images have found applications in various fields, from video games to forensic scenarios. 
Many applications have utilized GANs to create synthetic images that can be used for generating ground truth information used in training machine learning networks~\cite{deepsynth, han2023ensemble}.

There are few studies on applying GANs in the field of food image synthesis, such as RamenGAN \cite{ito2018food}, Multi-ingredient Pizza Image Generator (MPG) \cite{han2020mpg}, and CookGAN \cite{han2020cookgan}.
RamenGAN \cite{ito2018food} and its related work \cite{horita2019unseen} generate synthetic food images of noodles and rice based on a conditional GAN network.
MPG \cite{han2020mpg} uses conditional StyleGAN2 \cite{karras2020analyzing} to generate realistic pizza images with various ingredients.
CookGAN \cite{han2020cookgan} proposed a method to generate a meal image conditioned on a list of ingredients, with the types of foods limited to salad, a cookie, and a muffin.

In recent years, diffusion models have shown the ability to produce high-quality images while maintaining ease of use. 
The diffusion model defines a Markov chain to gradually add random noise to the data, and the model learns to reverse the diffusion process. 
As a result, the trained model is capable of constructing the desired data samples from the noise \cite{ho2020denoising}.
In \cite{dhariwal2021diffusion}, the authors demonstrate that the diffusion model outperforms GANs and achieves better image generation results on multiple datasets.
Additionally, in \cite{rombach2022high}, the authors propose latent diffusion models, which apply the diffusion process in the latent space of the data, resulting in a computational reduction and a boost in the visual fidelity of the generated images.
Despite diffusion models being applied to many downstream tasks \cite{pinaya2022brain, chen2023executing, luo2021diffusion}, the study of applying them to food image generation is still lacking.
In this paper, our work represents the first investigation and application of the diffusion model to food image generation.

\section{Method}
A block diagram of our proposed clustering-based conditional training strategy for the latent diffusion model, ClusDiff, is shown in Fig. \ref{fig:system_bd}. 
The system includes three parts: latent diffusion model training and inference, feature space food image clustering using affinity propagation, and clustering-based conditioning.
The motivation for this work is that food datasets usually exhibit high intra-class variance, with different cooking methods leading to variations within the same food class. 
As a result, training a generative model with simple class labels may not sufficiently cover the distribution in the training dataset.
In the training step of the generative model, we employ a clustering method to divide food images from a class label into several sub-classes based on their visual representations.
Subsequently, the generative model is trained conditionally on these sub-classes to further capture the variations within a food class.
When we use this trained generative model to generate food images, a food class label for conditional generation is represented as one of its sub-classes, determined by the frequency of occurrence in the training dataset. 
By doing this, the food images generated using ClusDiff better represent the distribution of the training dataset.

\subsection{Latent Diffusion Model}
\label{sec:diffusion}
There are two processes in diffusion models \cite{ho2020denoising}: (i) a forward diffusion process to diffuse an image into random noise, and (ii) a reverse diffusion process that converts the noise into the image from a learned data distribution during the iterative denoising process.  

In the forward diffusion process, a Markov chain is employed to gradually add noise to transform the training data into Gaussian noise. 
This discrete Markov chain is constructed with a predefined sequence of variance values $0<\beta_1, ..., \beta_T < 1$.
Given the input data distribution $q(x_0)$, a forward diffusing process at time $t$ is defined as Equation \ref{eq:forward_diff}:
\begin{equation}
    \begin{aligned}
    q(x_t|x_{t-1}) \sim \mathcal{N}\left(x_t; \sqrt{1 - \beta_t} x_{t-1}, \beta_t \mathbf{I}\right), \\
    q(x_t|x_0) \sim \mathcal{N}\left(x_t; \sqrt{\Bar{\alpha_t}} x_0, (1 - \Bar{\alpha_t}) \epsilon\right),
    \end{aligned}
    \label{eq:forward_diff}
\end{equation}
where $\epsilon \sim \mathcal{N}(0, \mathbf{I})$ denotes the injected noise, and the forward process can be interpreted as sampling $x_t$ at any time step $t$ in a closed form by using the notation: $\alpha_t := 1 - \beta_t$ and $\Bar{\alpha_t} := \prod_{n}^{s=1}\alpha_s$, which represent the noise level at each time step. 
At the final time step $T$, $x_T$ approximately becomes an isotropic Gaussian noise.  

In the reverse diffusion process, we can recover the input data distribution $q(x_0)$ from $p(x_T) \sim \mathcal{N}(x_T; 0, \mathbf{I})$. 
The equations for this process are defined as follows:
\begin{equation}
    \begin{aligned}
    p_{\theta}(x_{0:T}) \sim p(x_T) \prod_{t=1}^{T} p_{\theta}(x_{t-1}|x_t),\\
    p_{\theta}(x_{t-1}|x_t) \sim \mathcal{N}(x_{t-1}; \mu_{\theta}(x_t, t), \sum_{\theta}(x_t, t)).
    \end{aligned}
    \label{eq:reserve_diff}
\end{equation}


The training object of the diffusion model is to find a denoising autoencoder $\epsilon_\theta(x_t, t)$ to predict the injected noise $\epsilon$. The corresponding objective can be simplified as follows:

\begin{equation}
    \begin{aligned}
    L_{D} = \mathbb{E}_{t,x_0,\epsilon}\left[ ||\epsilon - \epsilon_\theta(x_t,t)||^2 \right],
    \end{aligned}
    \label{eq:diff_train}
\end{equation}
where $t$ is uniformly sampled from $\{1, ..., T\}$.

The latent diffusion\cite{rombach2022high} follows similar processes as the diffusion model, except that it uses the latent space of images as input samples to the diffusion model.
It utilizes a perceptual compression model\cite{esser2021taming} consisting of an encoder $\mathcal{E}$ and a decoder $D$. The encoder is used to map the image $x$ to the latent space $z=\epsilon(x)$ and the decoder is used to reconstruct the image $\bar{x} = D(z)$ from the latent space.
The latent space representation reduces the dimension of the input data, therefore focusing on the critical features of the input data and reducing the computational complexity of the diffusion model.

\subsection{Image Clustering using Affinity Propagation}
We utilize Affinity Propagation \cite{dueck2009affinity} to cluster the training images from one class into sub-classes. 
This helps in addressing the high intra-class variance problem of the food dataset.
A pre-trained ResNet-18 \cite{he2016deep}, fine-tuned on the Food-101 dataset for food classification, is used to map the images into feature vectors.
These feature vectors represent the visual features of the input food images and are used for clustering.
Affinity propagation creates clusters by sending messages between samples until convergence, and it does not require a pre-defined number of clusters.  
A small number of most representative samples are referred to as exemplars to describe the dataset.
Messages sent between pairs of samples indicate how well one sample can serve as the exemplar of the other, and these messages are iteratively updated in response to the values from other pairs. 
The updating process continues until convergence, and the final exemplars are used to form the clusters.

Suppose $z_i$ and $z_j$ are features vectors of two food images, we define $s(i,j)=1 - \frac{{z_i \cdot z_j}}{{\|z_i\| \cdot \|z_j\|}}$ as the similarity between $z_i$ and $z_j$, which is the cosine distance between two vectors.
In Affinity Propagation, two messages sent between the sample are responsibility $r$ and availability $a$.
Responsibility $r(i, j)$ describes how well $z_j$ to be the exemplar for $x_i$, relative to other candidate exemplars for $z_i$, and it is defined as:

\begin{equation}
    \begin{aligned}
    r(i, j) = s(i, j) - max [a(i, j') + s(i,j')\forall j'\neq j].
    \end{aligned}
    \label{eq:responsibility}
\end{equation}

Availability $a(i, j)$ describes suitability for $z_i$ to pick $z_j$ as its exemplar, considering other samples' preference for $z_j$ as an exemplar, and it is defined as:
\begin{equation}
    \begin{aligned}
    a(i, j) = min [0, a(j, j) +  \sum_{i'\notin\{i,k\}} r(i',j)].
    \end{aligned}
    \label{eq:availability}
\end{equation}

Both responsibility $r$ and availability $a$ are set to zero at the beginning and updated during each iteration until convergence.   
The damping factor $\lambda$ is used during the process to prevent numerical oscillations.
The updated functions for both responsibility and availability during each iteration are shown in Equation \ref{eq:APC_iter}:

\begin{equation}
    \begin{aligned}
    r_{n+1}(i,j) = \lambda \cdot r_n(i,j) + (1-\lambda) \cdot r_{n+1}(i,j),\\
    a_{n+1}(i,j) = \lambda \cdot a_n(i,j) + (1-\lambda) \cdot a_{n+1}(i,j),
    \end{aligned}
    \label{eq:APC_iter}
\end{equation}
where $n$ indicates the iteration step.

\begin{figure}
  \centering
     \includegraphics[width=0.98\linewidth]{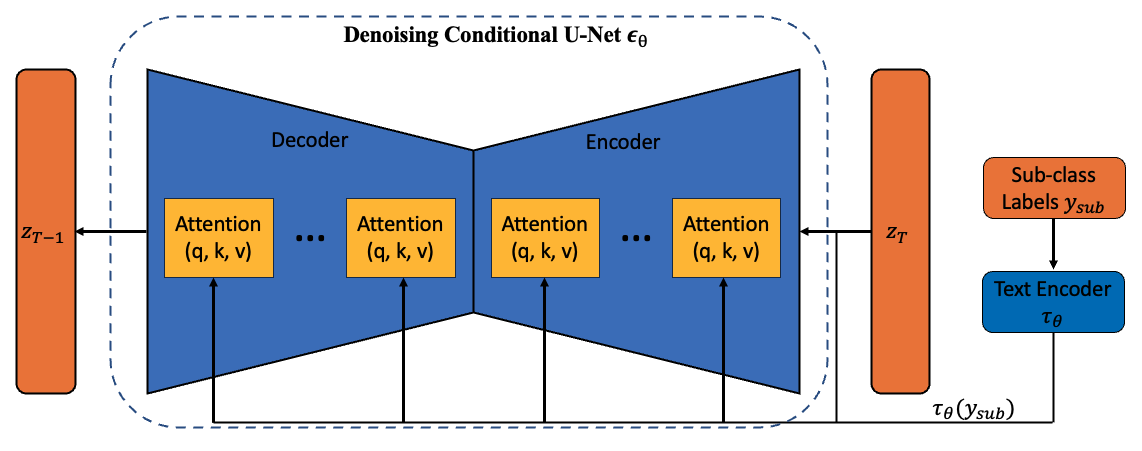}
  \caption{The block diagram of the denoising conditional U-Net architecture with cross-attention mechanism. Noted the attention layers (shown in green blocks) are mapped to each intermediate layer of U-Net.}
  \label{fig:unet}
\end{figure}

\subsection{Clustering-based Conditioning}
After the input food images are clustered into sub-classes, we denote the new class label of each food image by its food name followed by the cluster ID (e.g., burger\_1 for an image of a burger in cluster 1 of its class).  
A diffusion model is then trained to model the conditional distributions $p(x|y)$, where $x$ is the input image and $y$ is the class label of the input image.
To achieve this, we train a conditional denoising autoencoder $\epsilon_\theta(x_t, t, y)$, as mentioned in section \ref{sec:diffusion}, which allows us to control the image generation process through the class label $y$.
By conditioning the diffusion model on the new class label, we can generate food images that belong to a specific cluster of a food class, and this helps in generating more diverse food images within each food class.

Similar to the latent diffusion model\cite{rombach2022high}, we use a U-Net with an attention mechanism as our conditional denoising autoencoder. 
Figure \ref{fig:unet} shows how the sub-class labels $y_{sub}$ control the denoising process through U-Net. 
To incorporate the sub-class information into the denoising process, we use a pre-trained CLIP text encoder $\tau_\theta$ to project $y_{sub}$ into an embedding vector $\tau_\theta(y_{sub})$, which is then added to the intermediate layers of the U-Net using a cross-attention mechanism.
The implementation of the cross-attention mechanism is shown in the following equation:

\begin{equation}
    \begin{aligned}
    q = \pi_l(z_t)\cdot W_q^l, \quad k =& \tau_\theta(y_{sub})\cdot W_k^l, \quad v = \tau_\theta(y_{sub})\cdot W_v^l,\\
    \mathbf{A} =& softmax(\frac{qk^T}{\sqrt{d}}),
    \end{aligned}
    \label{eq:attention}
\end{equation}
where $q$, $k$, and $v$ are the query, key, and value for the attention mechanism, respectively. 
$l$ denotes the $l-th$ intermediate layer of U-Net, and $\pi_l(z_t)$ is the intermediate vector representation of U-Net.
$W_q$, $W_k$, and $W_v$ are learnable projection matrices, $\mathbf{A}$ is the attention map of cross-attention mechanism.

The new training objective of the conditional latent diffusion model is shown below:
\begin{equation}
    \begin{aligned}
    L_{LDM} = \mathbb{E}_{t,\mathcal{E}(z),y_{sub},\epsilon}\left[ ||\epsilon - \epsilon_\theta(z_t,t,\tau_\theta(y_{sub}))||^2 \right],
    \end{aligned}
    \label{eq:condtional_diff}
\end{equation}
where both denonising autoencoder $\epsilon_\theta$ and text encoder $\tau_\theta$ are jointly optimized together.

When generating images for a specific class of food, the latent diffusion model is conditioned with its sub-class labels to control the generation process, where the occurrence of the labels will follow the distribution of sub-classes within the class.

 \begin{figure*}
  \centering
     \includegraphics[width=0.98\linewidth]{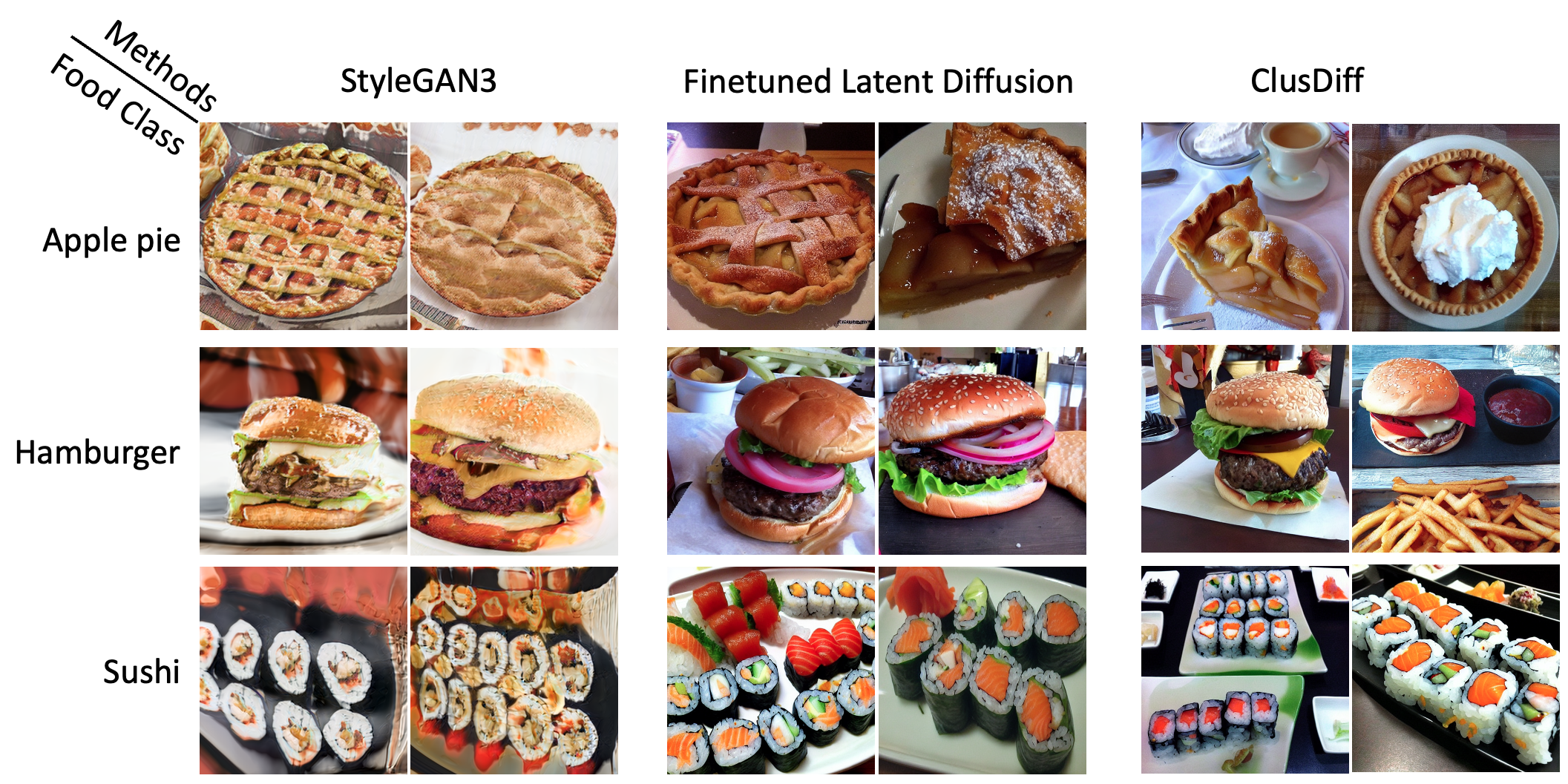}
  \caption{Examples of synthetic food images generated by each model. We manually selected three classes for demonstration, and the synthetic food images were randomly chosen. We can observe significant artifacts in the background of the food images generated by StyleGAN3. Furthermore, both diffusion-based methods, finetuned latent diffusion and ClusDiff, generate more distinct and refined details of the food images, such as the ingredients of the burger and the filling of the sushi.}
  \label{fig:gen_result}
\end{figure*}

\section{Experiments}
In this section, we first train a conditional StyleGAN3\cite{karras2021alias} on the Food-101\cite{bossard2014food} dataset as the GAN-based method. 
We then fine-tune a pre-trained stable diffusion model\cite{rombach2022high} on the Food-101 dataset as the baseline diffusion method.
Finally, we fine-tune a pre-trained stable diffusion with clustering-based conditioning as our proposed ClusDiff.
We evaluate ClusDiff and compare it to the GAN-based method and the baseline diffusion method using FID score \cite{heusel2017gans}.
Finally, we demonstrate the benefit of using generated synthetic food images as data augmentation by providing a case study to address the class-imbalance issue on the VFN-LT\cite{he2022long} dataset for the long-tailed food image classification.

\subsection{Dataset}
\subsubsection{Food-101}
Food-101\cite{bossard2014food} dataset consists of 101 food categories and each category contains 1000 images. 
The dataset contains a small amount of noisy data which are images with intense colors or wrong labels. 
All the images in Food-101 were rescaled to have a maximum side dimension of 512 pixels.
We further resize the images to 512 by 512 pixels for training generative models. 

\subsubsection{VFN-LT}
VFN-LT\cite{he2022long} dataset is the long-tailed version of the VFN dataset\cite{mao2020visual} constructed by removing data from each food class based on the food consumption frequency\cite{lin2021most} in the real world. 
Overall, the training set of VFN-LT is highly imbalanced containing 2.5K images from 74 common food classes with a maximum of 288 images per class and a minimum of 1 image per class. The test set of VFN-LT is kept balanced as 25 images per class for the long-tailed classification problem.

\subsection{Evaluation Metric}
\subsubsection{Frechet Inception Distance Metric (FID)}
FID\cite{heusel2017gans} is a metric that calculates the distance between the feature vectors of training and generated images, captured by a pre-trained Inception-v3\cite{szegedy2016rethinking} model. 
The FID measures the similarity between the features extracted from the training images and the generated images. 
A lower FID score indicates the feature distributions of generated images are close to real images, suggesting that the generator well captures the data distribution of the training images. 
A higher FID score indicates the generated images deviate from the real images, meaning the generator does not have a good performance.

\subsubsection{Long-tailed classification}
We use Top-1 image classification accuracy as the evaluation metric. In addition, we provide the performance on both head (instance-rich) and tail (instance-rare) classes. Following the benchmark protocol~\cite{he2022long}, there are 22 head classes and 52 tail classes in the VFN-LT dataset.

\subsection{Implementation Details}

\subsubsection{Food image generation}
The StyleGAN3 model is trained on the Food-101 dataset for 150 epochs with a batch size of 8, and the fixed learning rate is set to 0.002.
The reason we did not use pre-trained weights for StyleGAN3 is that the pre-trained models of StyleGAN3 have only been trained on face images, which does not contribute effectively to food image generation.
Both the baseline diffusion method and the clustering-based conditional diffusion method (ClusDiff) are fine-tuned on the pre-trained weight 'stable-diffusion-v1-4' for 100 epochs with a batch size of 4, and the fixed learning rate is set to 1e-5.
The pre-trained weight of the stable diffusion model is trained on a subset of the LAION-5B dataset \cite{schuhmann2022laion}, which consists of 5.85 billion general image-text pairs, including some food images. It is a common practice to finetune this pre-trained weight for downstream applications.

\subsubsection{Long-tailed classification}
 We use ResNet-18~\cite{he2016deep} pre-trained on ImageNet~\cite{IMAGENET1000} as the backbone and train the model for 150 epochs with batch size 128 using the SGD optimizer. The initial learning rate is set as 0.1 and decreased to 0.0001 using a cosine learning rate scheduler.

 \begin{table}[t]
    \centering
    \scalebox{1.2}{
    \begin{tabular}{cccc}
        \hline
        Datasets & \multicolumn{1}{c}{Food-101} \\
        \hline
        Methods & FID \\
        \hline
        StyleGAN3\cite{karras2021alias} &39.05  \\
        Finetuned Latent Diffusion \cite{rombach2022high}&30.39  \\
        \hline
        \textbf{ClusDiff} & \textbf{27.73} \\     
        \hline
    \end{tabular}
    }
    \caption{Food image generation results in terms of FID. The lower FID score indicates better diversity and fidelity of generated images and the best result is marked in bold. }
    \label{tab:foodgenresult}
\end{table}

\subsection{Image Generation Results on Food-101}
\textbf{Compared methods.} We compare our proposed ClusDiff with two other techniques: one of the latest GAN-based methods, StyleGAN3, and the baseline latent diffusion method. 
StyleGAN architectures have shown an ability to generate highly realistic images across multiple applications, including generating food images \cite{han2020mpg}. Therefore, we choose the latest variant, StyleGAN3 \cite{karras2021alias}, as the GAN-based method used for comparison. 
As for the baseline diffusion method, we select the latent diffusion model\cite{rombach2022high} to showcase the effectiveness of our proposed clustering-based training framework. 
This will allow us to demonstrate the improvements achieved by our modifications compared to the baseline.

We generate 100 synthetic images for each type of food in the Food-101 dataset using each image generation method and compare the FID scores. 
Fig. \ref{fig:gen_result} shows the sample results of synthetic food images generated by each method.
We manually selected three classes of food for demonstration, and the generated food images were randomly chosen. 
Table \ref{tab:foodgenresult} shows the comparison results for different food image generation methods.

From Fig. \ref{fig:gen_result}, we can see that compared to the synthetic food images generated by StyleGAN3, the images generated by the latent diffusion method show significant improvement in the background of food images.  
Furthermore, the latent diffusion method generates more refined details in the food region compared to StyleGAN3, as evident by the intricate textures of the sushi fillings and the distinct presentation of hamburger ingredients.
The FID scores from Table \ref{tab:foodgenresult} further confirm our visual inspections, showing that the baseline diffusion method outperforms StyleGAN3 by approximately 10 FID points.

As shown in Fig. \ref{fig:gen_result}, the visual difference between the synthetic images generated by the fine-tuned latent diffusion method and our proposed ClusDiff, is minimal. This is because ClusDiff maintains the original network structure of the latent diffusion model.

The comparison between ClusDiff and the baseline latent diffusion model is shown in Table \ref{tab:foodgenresult}. Since both methods are fine-tuned on the same pre-trained weight, this comparison demonstrates that ClusDiff significantly enhances the FID results of the latent diffusion model. 
This improvement indicates that our proposed modifications in ClusDiff effectively assist the latent diffusion model in capturing the underlying distribution from the training images, leading to better diversity and fidelity in the generated images.

\begin{table}[t]
    \centering
    \scalebox{1.}{
    \begin{tabular}{cccc}
        \hline
        Datasets & \multicolumn{3}{c}{\textbf{VFN-LT}} \\
        \hline
        Methods & Head (\%)  & Tail (\%)  & Overall (\%)  \\
        \hline
        Baseline &62.3 & 24.4 & 35.8  \\
        ROS\cite{ros} &61.7 & 24.9 & 35.9 \\
        RUS\cite{rus_ros} &54.6 & 26.3 & 34.8 \\
        CMO\cite{cmo} & 60.8 & 33.6 &42.1\\
        LDAM\cite{ldam} & 60.4 & 29.7 & 38.9 \\
        BS\cite{bsloss} & 61.3 & 32.9 & 41.9 \\
        IB\cite{ibloss} & 60.2 & 30.8 & 39.6 \\
        Focal\cite{focalloss} &60.1 & 28.3 & 37.8 \\
        Food2stage\cite{he2022long}  & 61.9 & 37.8 & 45.1 \\
        \hline
        \textbf{Baseline + ClusDiff} & \textbf{68.7} &\textbf{42.4} & \textbf{49.5}\\     
        \hline
    \end{tabular}
    }
    \caption{Top-1 accuracy on VFN-LT with tail class (Tail) and head class (Head) accuracy. The best results are marked in bold. }
    \label{tab:ltresult}
\end{table}

\subsection{Long-tailed Food Classification Results on VFN-LT}
We demonstrate the effective use of synthetic food images in addressing the challenges of class imbalance associated with the long-tailed classification problem. Specifically, we generate food images for each of the 74 food classes in the VFN-LT dataset to construct a balanced training set. The food images are generated by using a pre-trained stable diffusion model with our clustering-based conditioning on the Food-101 dataset for a fair evaluation. 

\textbf{Compared methods.} We compare with existing long-tailed image classification methods, including random over-sampling (ROS)\cite{ros}, random under-sampling (RUS)~\cite{rus_ros}, context-rich oversampling (CMO)\cite{cmo} and visual-aware hybrid sampling (Food2stage)\cite{he2022long} as data augmentation-based approaches. 
In addition, several loss functions have been commonly used to address class imbalance including label-distribution-aware margin loss (LDAM)~\cite{ldam}, balanced Softmax (BS)~\cite{bsloss}, influence-balanced loss (IB)~\cite{ibloss} and Focal loss~\cite{focalloss}. 
We use vanilla training as the baseline and our method as Baseline + ClusDiff by using a balanced training set containing original and synthetic food images generated by our proposed ClusDiff. 

Table~\ref{tab:ltresult} summarizes the results on the VFN-LT dataset. Our method achieves the best performance in terms of head, tail, and overall classification accuracy. 
Compared with existing data augmentation based approaches, using synthetic images is more effective, which improves the generalization ability and avoids the over-fitting issue for both instance-rich and instance-rare classes. Moreover, the use of synthetic data streamlines the training process, eliminating the need for any new loss functions, thus offering an advantage over existing methods and showing great potential for real-life applications.

\section{Conclusion and Future Work}
In this paper, we initially demonstrate the promising performance of latent diffusion methods in food image generation by comparing their results with one of the latest GAN-based methods, StyleGAN3.
We then propose a clustering-based conditional training framework, ClusDiff, to address the challenge of high intra-class variance within the food dataset.
This framework enables us to generate more diverse and representative food images that better reflect the underlying distribution of the training data.
Experimental results show that our proposed method outperforms the baseline diffusion model, showcasing its effectiveness in enhancing food image generation. Additionally, we conduct a thorough study on the utilization of synthetic food images to address the class-imbalance issue in long-tailed food classification. 

In the future, we plan to investigate integrating cluster information into the loss function or network structure of the diffusion model to better capture the variance of food images. This approach could potentially lead to further improvements in the quality and diversity of the generated food images.

\printbibliography


\end{document}